\documentclass[11pt,a4paper]{article}
\usepackage[hyperref]{emnlp_ijcnlp_2019}
\usepackage{times}
\usepackage{latexsym}

\usepackage{amsmath,amssymb}
\usepackage{url}
\usepackage{graphicx}
\usepackage{multirow}
\usepackage{multicol}
\usepackage{dblfloatfix}

\definecolor{bleudefrance}{rgb}{0.03, 0.27, 0.49}
\definecolor{carrotorange}{rgb}{0.91, 0.41, 0.17}

\aclfinalcopy 

\title{Building Task-Oriented Visual Dialog Systems Through Alternative Optimization Between Dialog Policy and Language Generation}

\author{Mingyang Zhou \qquad
  Josh Arnold \qquad
  Zhou Yu \\
  Department of Computer Science\\
  University of California, Davis\\
{\tt\{minzhou, jarnold, joyu\}@ucdavis.edu}
  }

\date{}

\begin{document}
\maketitle
\begin{abstract}
Reinforcement learning (RL) is an effective approach to learn an optimal dialog policy for task-oriented visual dialog systems. A common practice is to apply RL on a neural sequence-to-sequence (seq2seq) framework with the action space being the output vocabulary in the decoder. However, it is difficult to design a reward function that can achieve a balance between learning an effective policy and generating a natural dialog response. This paper proposes a novel framework that alternatively trains a RL policy for image guessing and a supervised seq2seq model to improve dialog generation quality. We evaluate our framework on the GuessWhich task and the framework achieves the state-of-the-art performance in both task completion and dialog quality.
\end{abstract}

\section{Introduction}

Visually-grounded conversational artificial intelligence (AI) is an important field that explores the extent intelligent systems are able to hold meaningful conversations regarding visual content. Visually-grounded conversational AI can be applied to a wide range of real-world tasks, including assisting blind people to navigate their surroundings, online recommendation systems, and analysing mass amounts of visual media through natural language. Current approaches to these tasks involve an end-to-end framework that maps the multi-modal context to a deep vector in order to decode a natural dialog response. This framework can be trained through supervised learning (SL) with the objective of maximizing the distribution of the response given a human-human dialog history. Given a large amount of conversational data, the neural end-to-end system can effectively learn to generate coherent and natural language.   

While much success has been achieved by applying neural sequence to sequence models to open visual grounding conversation, the visual dialog system also needs to learn an optimal strategy to efficiently accomplish an external goal through natural conversations. To address this issue, various image guessing tasks such us GuessWhich \cite{GuessWhich} and GuessWhat \cite{GuessWhat} are proposed to evaluate a visual-grounded conversational agent on its ability to retrieve visual content via conversing in natural language. To obtain an optimal dialog policy, reinforcement learning (RL) is introduced to enable the neural end-to-end framework to model a more effective action distribution by exploring different dialog strategies. With the application of RL, the visual dialog system can generate more consistent responses and achieve a higher level of engagement in the conversation when compared to a dialog system trained via SL \cite{visdialrl,GuessWhat_RL}. A typical way to apply RL on a dialog system is to assign a task-related reward to influence the utterance generation process by treating each output word as the action step. A significant limitation of this approach is that it is difficult to achieve an optimal dialog policy that can both effectively complete the external goal and generate natural utterances \cite{rethinking_action_space,visdialrl}. As there is no ground truth reference during the RL training stage, the dialog system can only leverage the reward signal when generating the response. However, this approach often deviates from natural language as it is challenging to define a comprehensive reward that considers all aspects of the dialog quality, and in addition, assigns appropriate rewards to the large word vocabulary action space.

In this paper we propose a novel learning curriculum to address the challenge of joint learning between the dialog policy and language generation for task-oriented dialog systems. In our framework, we separate the training of the image retrieval policy from dialog generation by applying RL, with the goal of achieving an optimal policy for guessing the target image at every turn. In addition, we apply a language model objective function to optimize the utterance generator to mitigate language degeneration. We specifically study this framework in the image guessing task, GuessWhich, where a conversational agent attempts to guess a target image by asking a series of questions. When compared to state-of-the-art RL visual dialog systems, our method achieves superior performance in both task-accomplishment and dialog quality.   


\section{Related Work}
\subsection{Visual Dialog System}
Visual dialog systems are an emerging area of interdisciplinary research that attracts both the vision and language communities due to the potential applications. \citet{visdial} proposed a visual dialog task in which a conversational agent attempts to answer questions regarding an assigned image based on a dialog history. To approach this task, they initially collected data by having two people chat about an image with one person acting as the questioner and the other as the answerer. GuessWhich \cite{GuessWhich} extends VisDial with the goal to build an agent that learns how to identify a target image through questions and answers. \cite{GuessWhat} additionally introduced a game in which a series of yes-or-no questions are asked by an agent in order to locate an object in an image. Many researchers approached these tasks via reinforcement learning (RL) with the goal of obtaining an optimal dialog policy. \citet{GuessWhat_RL}, for example, designed three rewards with respect to the goals of task achievement, efficiency, and question informativeness, in order to help the agent achieve an effective question generation policy for the GuessWhat game. \citet{visdialrl} applies reinforcement learning in the GuessWhich task and demonstrates a moderate improvement in accuracy compared to the supervised learning approach. Both methods apply RL on a neural end-to-end pipeline to jointly influence the language generation and dialog policy. Due the challenge of designing an appropriate reward for language generation, these methods generate responses that deviate from human natural language. \citet{JP}, proposed an approach involving hierarchical reinforcement learning and state-adaptation techniques that enable the agent to learn an optimal and efficient multi-modal policy. The bottleneck of \cite{JP}'s method, however, is that the system response is retrieved from a predefined human-written or system-generated utterance. The number of predefined responses are limited, therefore,  this method does not easily generalize to other tasks in real-world settings. We address these limitations by applying RL on a reduced, yet more relevant action space, while optimizing the dialog generator in a supervised fashion. We alternatively optimize policy learning to language generation to combine the two tasks together.

\begin{figure*}[h!]
\centering
\includegraphics[width=16cm]{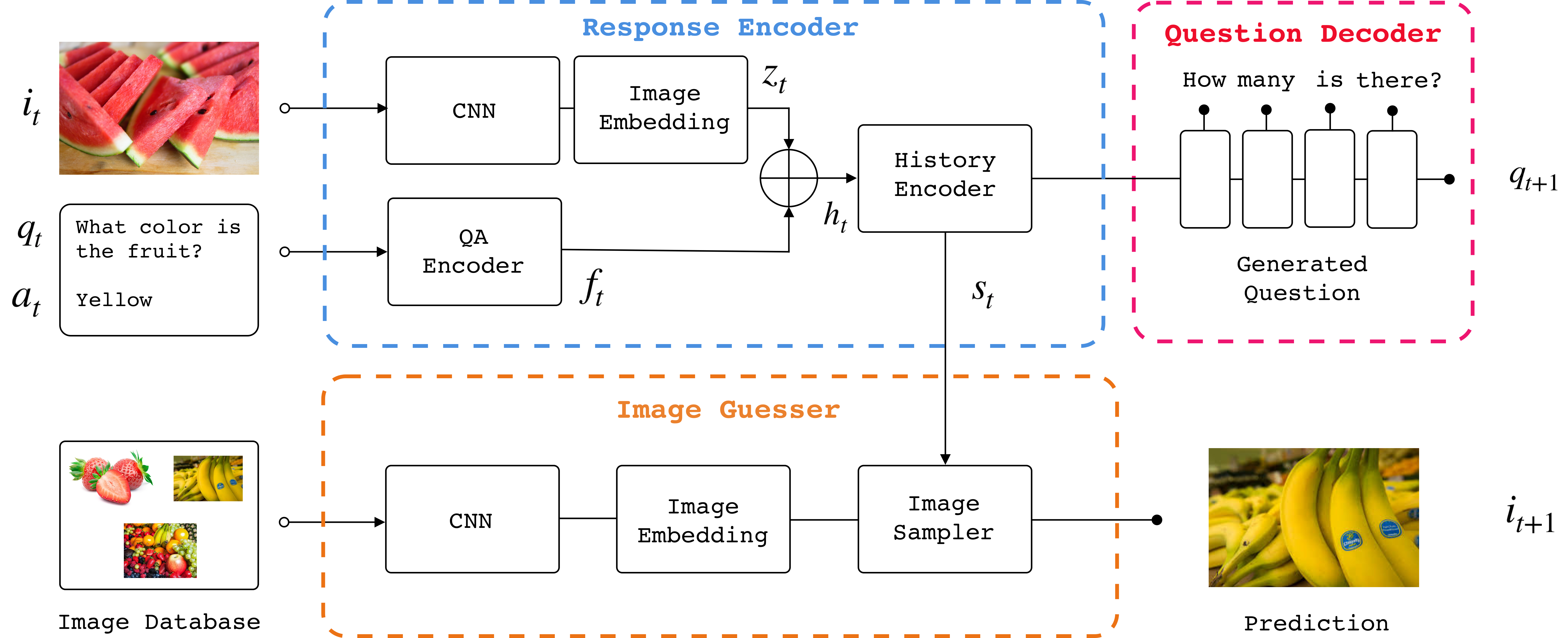}
\caption{The proposed end-to-end framework of the conversation agent for GuessWhich task-oriented visual dialog task}
\end{figure*}

\subsection{RL on Task-oriented Dialog System}
Various RL-based models have been proposed to train task-oriented dialog systems \cite{Williams_2007}. 
In order to build a traditional modular-based dialog system, researchers first identify the semantic representation, such as the dialog acts and slots in user utterances. Then they accumulate these semantic representations over time to track the dialog state. Finally they apply RL to learn an optimized dialog policy given the dialog state \cite{dialogact_RL,sentimen_analysis_dialogact}. 
Such modular-based dialog systems are effective in narrow task domains, such as searching a bus route schedule or reserving a restaurant through natural language, but they fail to generalize to complex settings where the size of the action space increases. Owing to the development of deep learning, RL on neural sequence-to-sequence models has been explored in more complex dialog domains such as open-domain conversation \cite{Deep_RL_Dialog} and negotiation \cite{dealNodeal}. However, due to the difficulty of assigning appropriate rewards when operating in a large action space, these frameworks cannot generate fluent dialog utterances. \citet{rethinking_action_space} proposed a novel latent action RL framework to marry the advantage of a module based approach and sequence-to-sequence approach. They learned the optimal dialog policy in a complex task-oriented dialog domain while achieving decent conversation quality. Here, we study a similar issue in a multi-modal task-oriented dialog scenario. We propose an iterative approach using RL to optimize the dialog policy and SL to optimize the generation of the system response.

\section{Method}
\subsection{Problem Setting}
In the GuessWhich problem, we aim to build an agent (Q-Bot) that attempts to guess an image $i_{tgt}$ that another agent (A-Bot) knows by asking it a series of questions. At the beginning of the conversation, the Q-Bot is primed with a short caption $c$ of the target image that is only known by A-Bot. At every round $t$, the Q-Bot generates a question $q_t$ to elicit as much information as possible about the target image and the A-Bot provides an appropriate answer $i_t$ with regard to $q_t$ and the target image. In the end, the agent guesses the target image among a set of images considering the entire conversation.

In addition, our dialog system also guesses a candidate image $i_t$ out of an image database $\mathcal{I} = \{i_k\}_{k=0}^{m} $ at every turn. This action models the process of sequentially updating the visual belief state on the target image based on the latest dialog history. Conditioned on the current guessed image and the prior dialog contexts, the system will generate an optimal question in order to get the maximum information from A-Bot that can strengthen the system's belief on the target image. At the end of the conversation, our Q-Bot will guess the target image based on the multimodal contexts $s_n = (q_{1:n}, a_{1:n}, i_{1:n}, c)$ consisting of the dialog history and the trajectory of guessed images.

\subsection{Model Architecture}
Our Q-Bot is constructed on top of a hierarchical encoder-decoder framework \cite{Hre_Enc_Dec}, which consists of three major components: The \textbf{Response Encoder}, the \textbf{Question Decoder}, and the \textbf{Image Guesser}. We introduce each component as follows:
\paragraph{Response Encoder}
The goal of the response encoder is to append the question $q_{t}$, the answer $a_{t}$, and the guessed image $i_{t}$ received at current round to the dialog history and obtain an updated vector representation of the multimodal context $s_{t}$. The image $i_t$ is encoded with a pre-trained convolutional neural network VGG-16 \cite{vgg} followed by a linear embedding layer and the image feature vector denoted as $z_{t}$. For the question and answer pair at the current round $(q_{t}, a_{t})$, we map them to a hidden state vector $f_{t}$ through the LSTM based \textit{QA Encoder}. We then apply a linear projection on the concatenation of $f_{t}$ and $z_{t}$ in order to obtain the multi-modal context vector $h_t$ for the current round. The context vector is then passed through another LSTM encoder: \textit{History Encoder} generates an updated dialog history representation $s_t = \text{HistoryEnc}(h_t, s_{t-1})$. We denote the trainable parameters for Response Encoder as $\theta_{e}$.    
\paragraph{Question Decoder} 
The question decoder is a two-layer LSTM network initialized with the most updated dialog history representation vector $s_t$ from the response encoder. It will sequentially sample the words to come up with the next question $q_t$. The learned parameters for question decoder are denoted as $\theta_{d}$.
\paragraph{Image Guesser}
 The Image Guesser attempts to identify the candidate image that best aligns with the dialog history. Given an image database $\mathcal{I} = \{i_k\}_{k=0}^{m}$ where we sample the candidate image, we first extract the image feature representations $\{z_k\}_{k=0}^{m}$ for all candidate images with the convolutional neural network and image embedding layer defined in response encoder. Then, we can sample a candidate image $i_k$ for the current turn based on the euclidean distance $d(z_k, s_t)$ between the image feature of the candidate image and the current dialog history vector. The image with the smallest euclidean distance is selected as the guess $i_t$ at the current round. The associated parameters for image guesser are defined as $\theta_{g}$.
 \subsection{Training Dialog System}
 We follow a two-stage training fashion as introduced in many previous end-to-end RL dialog systems \cite{visdialrl, GuessWhat_RL, rethinking_action_space}, where we first pre-train the dialog framework with a supervised objective then apply reinforcement learning to learn an optimal policy to retrieve the target image. The Supervised pre-training is a critical step that facilitates an effective policy exploration for RL training, as it is difficult to explore a complex action space with limited prior knowledge. During RL training, we introduce an alternative learning method between dialog policy exploration and natural utterance generation that addresses the issue of language degeneration in previous RL based visual dialog systems \cite{visdialrl}. We introduce each training method as follows.    
 \subsubsection{Supervised Pre-training}
 During the supervised pre-training process, we jointly optimize the objective to generate questions and also predict target image features from dialog contexts. The task of question generation is optimized by maximizing the log conditional probability of the next question dependent on a ground truth dialog for every round of the conversation. For the image feature prediction, we minimize the mean square error (MSE) between the target image feature $z_{tgt}$ and the dialog context vector $s_t$ at each round. The joint loss function for supervised pre-training is:
 \begin{multline}
     \mathcal{L}_{SL}(\theta_r, \theta_d, \theta_g) = \alpha \sum_{t=0}^{n}\log{p(q_t|s_t)} \\
     + \beta \sum_{t=0}^{n} \text{MSE}(z_{tgt}, s_t)
     \label{eq:1}
 \end{multline}
 Where $\alpha$ and $\beta$ are weights assigned to the objective function of each task in the joint objective function. With SL pre-training process, the dialog system is facilitated with the ability to estimate a visual object and emit a natural language sentence given a dialog context.
 
 \subsubsection{Reinforcement Learning on Image Retrieval}
 In our framework, we treat the sequence of image guess through out the conversation as a partially observable markov decision process and train a policy network via RL to obtain an optimal strategy to retrieve the target image. We formally describe state, policy, action, rewards, and the training procedures in our pipeline. 
 \paragraph{State}
 The dialog states in our framework consist of a combination of multimodal contexts, including the image caption $c$, the dialog history with A-Bot $[q_1, a_2, \dots, q_t, a_t]$, and the image guessing trajectories $[i_1, i_2, \dots, i_t]$.  
 \paragraph{Policy}
 The dialog policy $\pi_{\theta_r, \theta_g}(i_t|S_t)$ is a stochastic policy that samples the candidate image to guess from an image set based on the previous dialog histories. The policy is learned from response encoder and image generator which is parameterized via $\theta_r$ and $\theta_g$. 
 \paragraph{Action}
 The full action space is the number of images in the database that we can sample to guess an image. As the pre-trained process already enables the system to approximate a target image feature $z_{tgt}$ with the dialog history representation vector $s_t$,  we reduce the action space to the top-K nearest images, $s_t$, based upon the euclidean distance. The probability to sample an image $i_j$ is gained by applying a softmax function over the top-K candidates on their distance to $s_t$: $\pi(j) = \frac{e^{-d_j}}{\sum_{k=1}^{K} e^{-d_k}}$. $d_j$ represents the mean-square-distance between the $j$-th image and the dialog history state vector $s_t$.
 \paragraph{Rewards}
We use the ranking percentile of the target image with respect to the dialog history vector $s_t$ as the reward signal to credit the guess at each turn. The goal is to maximize the expectation value of the discounted return $\mathbb{E}[\sum_{t = 1}^{n} \gamma^t r_t]$ over the n-round conversation. $r_t$ is the ranking percentile of target image at round $t$ and $\gamma$ is the discounted factor between $(0,1)$. 
 \paragraph{Training Procedure}
 Inspired from the RL training process on the iterative image retrieval framework \cite{dialog_interactive_image_retrieval}, we apply the policy improvement theory \cite{Intro_RL} to estimate an improved policy $\pi^{*}(s_t)$ from an existing policy $\pi(s_t)$ obtained from the pre-trained dialog system. Given a dialog state $s_t$ and the action $a_t$ derived from the existing policy, the value estimated by the current policy for taking the action $i_t$ is $Q_{\pi}(s_t, i_t) = \mathbb{E}[\sum_{t^{'} = t}^{n} \gamma^t r_t]$. To improve this, we explore a different action $i_t^{*} \neq i_t$ such that a larger policy value $Q_{\pi}(s_t, i_t^{*}) >  Q_{\pi}(s_t, i_t)$ estimated with the current policy is achieved. Then we can adjust the existing policy $\pi(s_t)$ to a new policy $\pi^{*}(s_t)$ that executes that optimal action $i_t^{*}$ given the current dialog state. The parameters of the policy can be effectively optimized via a cross entropy loss function conditioned on the derived optimal action $i_t^{*}$:
 \begin{equation}
     \mathcal{L}_{RL}(\theta_r, \theta_g) = \mathbb{E}[-\sum_{t = 1}^{n} \log(\pi_{\theta_r, \theta_g}(i_t^{*}|s_t))] 
     \label{eq:2}
 \end{equation}
 Compared to the previous RL visual-grounded conversational agent, \cite{visdialrl}, there are several advantages of conducting policy learning on the action level of guessing the image. First, the action space of the top-k nearest neighbors are much smaller compared to the vocabulary size of the output words which reduces the difficulty to explore optimal strategies. Second, only the parameters of response encoder and image guesser will be optimized during the RL training stage. The question decoder stays intact so that it is less likely for the dialog system to suffer from language deviation. 
 \subsubsection{Alternating Policy Learning and Language Generation}
 Although the parameters of the decoder won't be impacted during the RL training stage, the shared response encoder of the dialog context is still optimized with policy learning. The language distribution captured by both the response encoder and question decoder will gradually be differentiated from the original human dialog distribution. To prevent the potential language degeneration behavior, we alternatively optimize the dialog system with a policy learning objective in equation \ref{eq:2} and the language model objective function in equation \ref{eq:1} at every other epoch. It assures the dialog system maintains a good estimation of the human language distribution while also effectively exploring various dialog actions in order to achieve the task of guessing the right image.

\begin{table*}[]
\resizebox{\textwidth}{!}{%
\begin{tabular}{llll}
\hline Image & RL-Q-IG & SL-Q-IG & RL-Q \\ \hline \\ [-1.5ex]

\multirow{1}{*}{
\includegraphics[width=3cm]{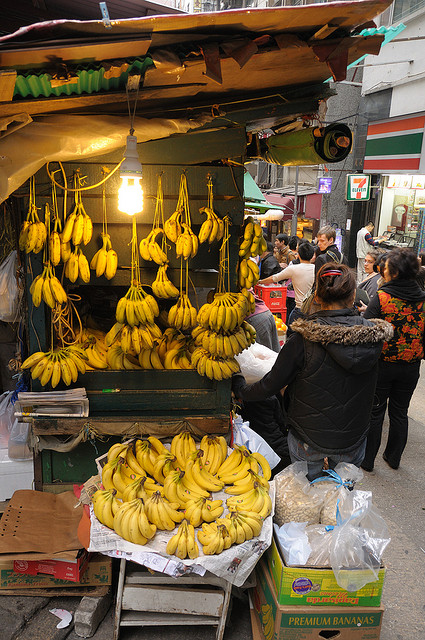}
}

& \color{bleudefrance} Q: are bananas in bunch? & \color{carrotorange} Q: what color are planes? & Q: is this outside? \\
& A: yes it's in bunch & A: no planes there & A: yes it is \\ 
& Q: are they on table? & \color{carrotorange} Q: are they in city? & Q: is there any people? \\ 
& A: yes they are & A: yes it is & A: yes ... \\ 
& \color{black} Q: is this in kitchen? & \color{carrotorange} Q: are they in baskets? & \color{carrotorange} Q: is this in zoo? \\ 
& A: I'm not sure & A: yes they are & A: no it is not \\
& Q: any people? & Q: any people? & Q: are there any people? \\ 
& A: Yes there are several people & A: yes there are & A: yes there are \\ 
& \color{bleudefrance} Q: any other fruits? & \color{carrotorange} Q: animals? & Q: is there any people? \\ 
& A: NO, only bananas & A: no animals & A: yes ... \\ 

\end{tabular}}
\caption{A dialog example with the ground truth caption: \textbf{bunches of bananas hang on a wall and arranged for sale.}  \color{bleudefrance} blue \color{black} indicates ideal relevant questions and \color{carrotorange} orange \color{black} indicates less relevant questions.}
\label{tab:my_label}
\end{table*}

\section{Experiments}
\subsection{AI-AI Image Guessing Game}
We evaluate the performance of our task-oriented dialog system by playing the image guessing game, GuessWhich, with an automatic answer bot. Our conversational agent's goal is to locate the target image out of the 9,628 test images by interacting with the other player in five conversation exchanges. We evaluate agent on both goal achievement and utterance generation quality using two automatic evaluation metrics Percentile Mean Rank (PMR) and perplexity respectively. PMR estimates how good the agent can rank the target image against other candidates in the test database based on its current dialog state. Perplexity estimates the closeness of the generated response to a reference utterance given a dialog context from the VisDial dataset.

\subsection{Human-AI Image Guessing Game}
To evaluate the ability of our task-oriented dialog system in a realistic conversational scenario, we also make our agent play the image guessing game with human users. The games are set up as 20-image guessing games where the agent attempts to guess a target image outside of a pool of 20 candidate images by asking a human player 5 rounds of questions. The objective of the human player is to play the role of answer bot and answer agent's question with respect to the target image. 

In this setting, the performance of the agent on task accomplishment is evaluated by the game win rates. The quality of the dialogs are manually rated on four criteria: fluency, comprehension, diversity and relevance. Fluency defines the naturalness and readability of the generated question in English. Comprehension represents the consistency of the generated question with respect to the previous dialog context. Diversity evaluates the uniqueness of the questions generated within one game. Relevance presents how well the asked question is related to the target image and the given caption.  

\subsection{Comparative Models}
We compare the performance of our model with state-of-the-art task-oriented visual dialog systems. Meanwhile we also perform an ablation study to evaluate the contribution of different designs in our framework. We introduce each model as follows:

\textbf{SL-Q}: The dialog agent from \cite{visdialrl}, which is trained with a joint supervised learning objective function for language generation and image prediction. 

\textbf{RL-Q}: The dialog agent from \cite{visdialrl} which is fine-tuned on a trained SL-Q by applying RL to the action space of output word vocabulary. 

\textbf{SL-Q-IG}: The dialog agent from this framework is build on top of the SL-Q. Compared to SL-Q, SL-Q-IG has an additional image guesser module that makes a guess on target image at every round. SL-Q-IG also has an image encoder which fuses the guessed candidate image into the dialog history tracker. We only train this model with the supervised learning objective introduced equation \ref{eq:1}.

\textbf{RL-Q-IG}: We use RL method to fine-tune SL-Q-IG. The RL method used is applied on action space of guessing candidate image. We alternate the model to optimize towards dialog policy learning and language generation.

\textbf{RL-Q-IG-NA}:  We fine-tune SL-Q-IG by applying RL to the action space of guessing candidate image and only optimized with policy learning objective function alone.

\textbf{RL-Q-IG-W}: The dialog agent from our framework, which is fine-tuned on a trained SL-Q-IG by applying reinforcement learning on output word vocabulary. It follows the same training procedures as RL-Q to conduct policy learning.

All the SL dialog agents are trained on the VisDial Dataset with the default setting from \cite{visdialrl} for 40 epochs. The RL dialog agents are then fine-tuned on their corresponding SL dialog agents for another 20 epochs. We evaluate every model on AI-AI image guessing games with the same answer bot, trained on the Visdial Dataset with the objective of visual question answering. We only evaluate RL-Q, SL-Q-IG and RL-Q-IG in human evaluation. 

\subsection{Human-AI Evaluation Implementation}\label{human eval setting}
In order to evaluate the effectiveness of the model, we designed three human evaluation tasks. Six college students were recruited to conduct the evaluation. Each student evaluated 100 games using the ground truth captions and 30 games using human generated captions. An additional three evaluators each completed 30 rounds of the relevancy experiment.

\textbf{Ground Truth Captions} We generated 100 image guessing games that used the ground truth captions to ensure a consistent amount of information is supplied across all human evaluators. Each game consists of a randomly selected set of 20 images from the VisDial Dataset, with one image randomly chosen as the target. For each game, we test three different models, each twice, resulting in a total of 600 evaluated games from the 100 generated games. We keep the identity of the models anonymous to the evaluator.

During each game, the human evaluator is presented with a target image the agent is trying to guess. Five rounds of Q\&A take place in which the agent asks a question to elicit information and the human evaluator responds with a relevant truthful answer. At the end of each game, the evaluator is asked to rate the conversation on four criteria: fluency, relevance, comprehension and diversity.

\textbf{Human Generated Captions} In order to distinguish SL-Q-IG and RL-Q-IG in a more natural setting, we generate an additional 30 games, similar to the previous human evaluation task, except when beginning the game, the evaluator is asked to provide the caption for the target image instead of using the ground truth.

\textbf{Relevance Experiment} We noticed that the human evaluators found rating dialogues on the relevance criteria challenging and nuanced. In order to reduce the difficulty of rating dialogues using the relevance criteria, we designed a separate experiment in which, using the conversations obtained from the previous 600 evaluated ground truth games, a human evaluator is presented with three complete conversations side by side at each round. The evaluator then selects the most relevant conversation out of the three that corresponds to the image caption. Each of the three conversations have the same caption, however, correspond to a different model, thus allowing for an effective comparison between the relevancy of each model.

\section{Results}
\subsection{Results on AI-AI Image Guess Game}
\begin{figure}[h!]
\centering
\includegraphics[width=8cm]{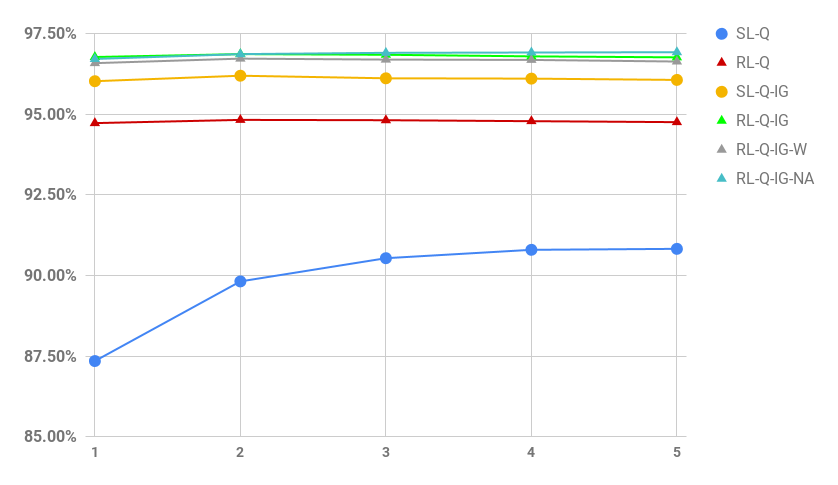}
\caption{The percentile mean rank (PMR) over the 5-round dialog in the AI-AI image guessing game}
\label{fig:2}
\end{figure}

\paragraph{Image Retrieval}
It is clear from Fig~\ref{fig:2} that our dialog system significantly outperforms the baseline models from \cite{visdialrl} in terms of PMR on every round of the dialog. PMR estimates how good the agent can rank the target image against other candidates in the test database. The biggest improvement gap is observed between SL-Q-IG and SL-Q. In comparison to SL-Q, SL-Q-IG tracks the additional context from the previously guessed images which leads to a better estimation of the target image. RL-Q-IG has better performance compared to SL-Q-IG in terms of PMR. This suggests that fine-tuning dialog systems with RL can further improve the success of guessing the correct image. The best image retrieval result is achieved by RL-Q-IG-NA, as the objective function of RL-Q-IG-NA is based solely on policy learning without consideration for the dialog generation quality.

\begin{table}[h!]
\small
\centering
\begin{tabular}{|l|l|l|}
  \hline
  Model & PMR & Perplexity \\
  \hline
  SL-Q & $90.07\%$ & $79.49$ \\
  SL-Q-IG & $96.09\%$ & $61.42$ \\
  \hline
  RL-Q & $94.78\%$ & $544.97$\\
  RL-Q-IG & $96.81\%$ & $\boldsymbol{54.66}$ \\
  RL-Q-IG-NA & $\boldsymbol{96.88\%}$ & $363.88$\\
  RL-Q-IG-W & $96.65\%$& $227.35$\\
  \hline
\end{tabular}
\caption{RL-Q-IG-NA performs best in PMR and RL-Q-IG perform best in perplexity}
\label{table:table_2}
\end{table}
\begin{table*}[t]
  \centering
  \small
  \begin{tabular}{|c|c|c|c|c|c|}
    \hline
     Model & Win & Fluency & Relevance & Comprehension & Diversity \\
     \hline
     RL-Q & 59.6 & 4.19 & 3.22 & 2.60 & 2.50 \\
     \hline
     SL-Q-IG & 62.7 & 4.18 & 3.96 & 3.18 & 3.22 \\
     \hline
     \textbf{RL-Q-IG} & \textbf{67.5} & \textbf{4.40} & \textbf{4.02} & \textbf{3.50} & \textbf{3.25} \\
     \hline
  \end{tabular}
  \caption{Evaluation results on the human-AI image guessing game initialized with ground truth captions}
  \label{tab:1}
\end{table*}

\begin{table*}[t]
  \centering
  \small
  \begin{tabular}{|c|c|c|c|c|c|}
    \hline
     Model & Win & Fluency & Relevance & Comprehension & Diversity \\
     \hline
     RL-Q & 29.2 & 4.04 & 2.88 & 2.71 & 2.29 \\
     \hline
     SL-Q-IG & 40.6 & 4.16 & 3.19 & 2.75 & 2.69 \\
     \hline
     \textbf{RL-Q-IG} & \textbf{67.6} & \textbf{4.23} & \textbf{3.74} & \textbf{3.32} & \textbf{3.06} \\
     \hline
  \end{tabular}
  \caption{Evaluation results on the human-AI image guessing game initialized with human generated captions}
  \label{tab:2}
\end{table*}

Although our framework achieved an improved image retrieval accuracy, we observed, however, that there is little improvement gained in PMR after additional rounds of conversation. We suspect this is partially due to the fact that images from MSCOCO are composed of a diverse selection objects and background scenes, thus making images easily distinguishable with a detailed caption. In cases where candidate images are visually similar or the given caption is not informative, additional rounds of dialog are necessary to identify the target image.

\paragraph{Language Generation}
We observe a marginal increase of perplexity from SL-Q to RL-Q in Table~\ref{table:table_2}, thus demonstrating that there is a bottleneck when applying RL to improve the language generation. By decoupling the policy learning from the language generation and alternatively optimizing the dialog policy and language model, our RL-Q-IG avoids language deviation while still achieving an optimal dialog policy for the image retrieval task. To further evaluate the contribution from the RL and alternative training curriculum, we conduct two ablation studies. RL-Q-IG-NA is fine-tuned with a policy learning objective that excludes alternatively applying the language model loss. While RL-Q-IG-NA only achieves an incremental improvement over the full framework RL-Q-IG in terms of the PMR rate with less than $0.1\%$, it suffers from a dramatic increase of perplexity from 61.42 to 363.88, thus suggesting that alternatively applying the supervised learning objective can prevent the language model from deviating from the human language distribution. We additionally apply policy learning on the question decoder of SL-Q-IG and follow the RL fine-tuning process in \cite{visdialrl} to train the agent, RL-Q-IG-W. While applying word-level RL enables RL-Q to achieve a moderate improvement over SL-Q in terms of PMR, we did not observe, the same degree of advantage in RL-Q-IG-W over SL-Q-IG. Additionally, RL-Q-IG-W is affected by a marginal increase in perplexity in comparison to the SL pre-trained agent, which approves the drawbacks of applying RL on a large action space in language generation.

\subsection{Results on Human-AI Image Guess Game}
The performance of a dialog agent evaluated with a user simulator does not necessarily reflect its performance on real human \cite{GuessWhat}. We conduct human evaluation on different dialog agents. From the results summarized in Table~\ref{tab:1} and Table~\ref{tab:2}, we observe a consistent optimal performance of our method from conversations with AI agent to conversations with real human. Our RL-Q-IG significantly outperforms the baseline RL agent in all criteria for both settings. RL-Q-IG's advantage over SL-Q-IG is not significant in the game when agents are primed with ground truth image caption. This observation correlates with the result in the Human-AI game, as both RL-Q-IG and SL-Q-IG achieve superior PMR over $96\%$ when presented with the ground truth caption. However, if a human generated caption is given, the performance of the SL pre-trained agent suffers a big drop in all metrics except fluency while our RL agent maintains similar performance. Applying RL to fine-tune the dialog system  enables the agent to generate more consistent dialogs in unseen scenarios. We also notice a degradation of the baseline RL agent from its performance with the user simulator, which suggests deviation from natural language is due to the sub-optimal RL training on a large action space.

Besides a marginal improvement over the RL baseline model and SL pretrained agent in terms of decreased repetition and grammar mistakes, there is a distinct superiority in regards to the relevance to the image caption in the questions generated from our RL agent. For example, in Table~\ref{tab:my_label}, we demonstrate the three dialogs generated by RL-Q-IG, SL-Q-IG and RL-Q on one game. Given the image caption \textit{bunches of bananas hang on a wall and arranged for sale.}, RL-Q and SL-Q-IG ask very general questions that are not related to the caption such as \textit{``planes"}, \textit{``zoo"} and \textit{``animals"}. In comparison, our agent asks high-quality questions regarding the caption that covers \textit{``bananas"} and \textit{``fruits"}. These questions help our RL agent obtain useful information to guess the target image. This advantage is also evident from the results of comparative evaluation on the degree of relevance of the questions in Table \ref{tab:relevance}. We credit the positive result to the dialog policy, which explores multiple paths to conduct the conversation. The optimal path will involve a set of questions that obtains the maximum information of the target image such that it can construct the best estimation of the target image. 

\begin{table}[]
    \small
    \centering
    \begin{tabular}{|c|c|}
        \hline
         Model & Prefered (\%) \\
         \hline
         RL-Q & 8.93\\
         \hline
         SL-Q-IG & 39.90 \\
         \hline
         \textbf{RL-Q-IG} & \textbf{51.20} \\
         \hline
    \end{tabular}
    \caption{Results on comparative evaluation of relevance on the human-AI image guessing dialogs}
    \label{tab:relevance}
\end{table}

\section{Conclusion and Future Work}
We present a novel framework for building a task-oriented visual dialog system. We model the agent to simultaneously optimize two actions: guessing the image and generating effective questions. We achieve this simultaneous optimization through alternatively applying reinforcement learning to obtain an effective image guessing policy, whilst also applying supervised learning to enhance the quality of generated questions. By decoupling the policy learning from language generation, we overcome language degeneration in the word-level reinforcement learning framework. Both analytical and human evaluation suggests our proposed framework leads to a higher task completion rate and an improved dialog quality.

In the future, we plan to collect a fashion retrieval visual dialog dataset which simulates a realistic application for multi-modal dialog systems. To address the limitation of a high image retrieval rate with just the use of captions from the VisDial dataset, we plan to format a challenging candidate image pool in which images are visually similar to each other. This will incentivize the dialog system to conduct multiple rounds of dialog in order to retrieve the target image successfully. Furthermore, we will explore additional task-oriented settings where we can decouple task accomplishment from language generation to evaluate the extent our framework can generalize to other conversational tasks.  

\section*{Acknowledgments}
This work was supported in part by Cisco. 




\bibliography{emnlp_ijcnlp_2019}
\bibliographystyle{acl_natbib}
\begin{figure*}[!b]
\centering
\includegraphics[width=15cm]{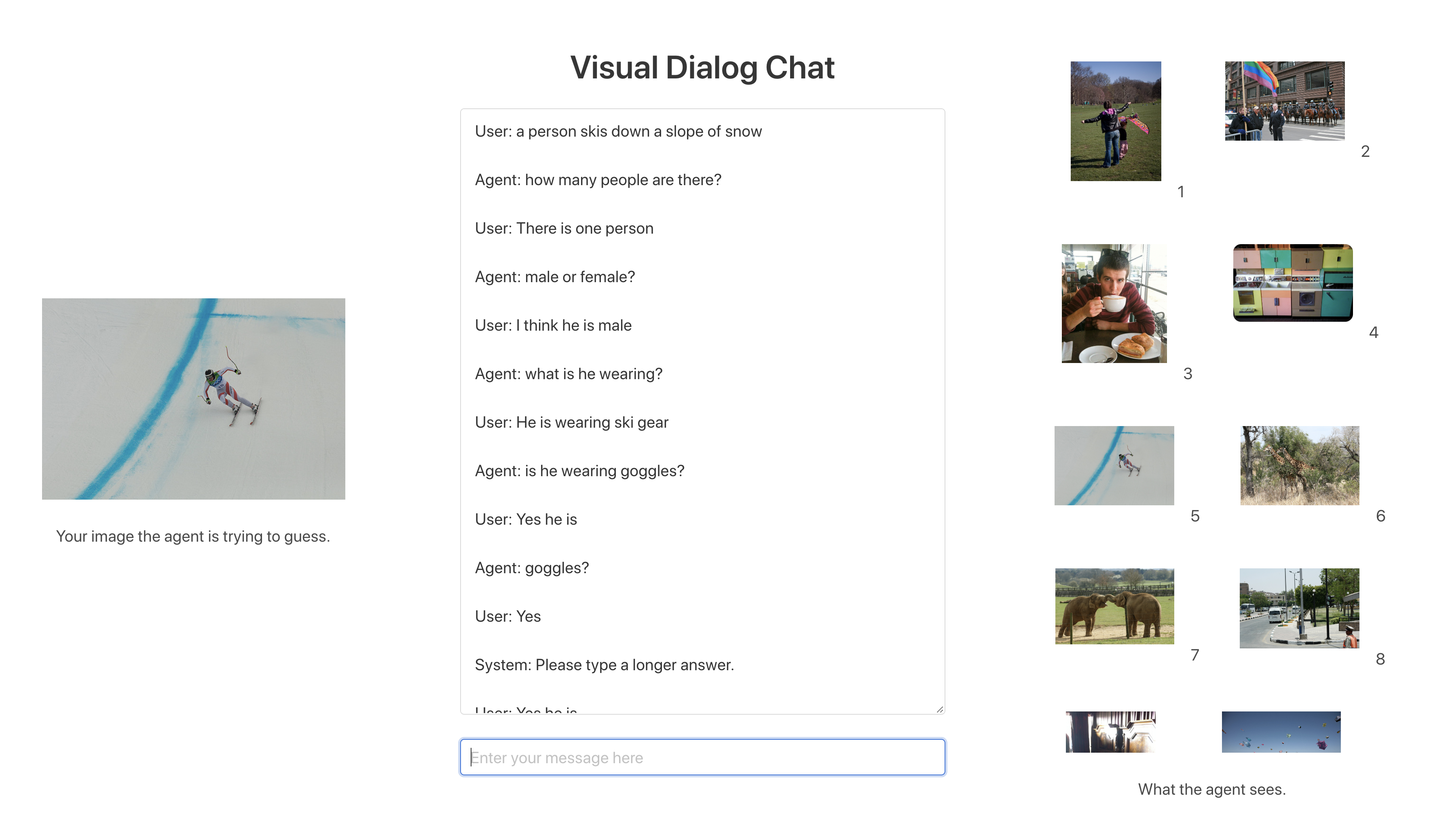}
\caption{The web interface for human-AI guessing game. The left image is a target image randomly sampled from \cite{visdial}. The center section is a chat platform for human to communicate with a trained Q-Bot. On the right hand side are the 20 candidate images sampled for the Q-Bot to retrieve the target image.}
\label{fig:3}
\end{figure*}

\begin{figure*}[!b]
\centering
\includegraphics[width=15cm]{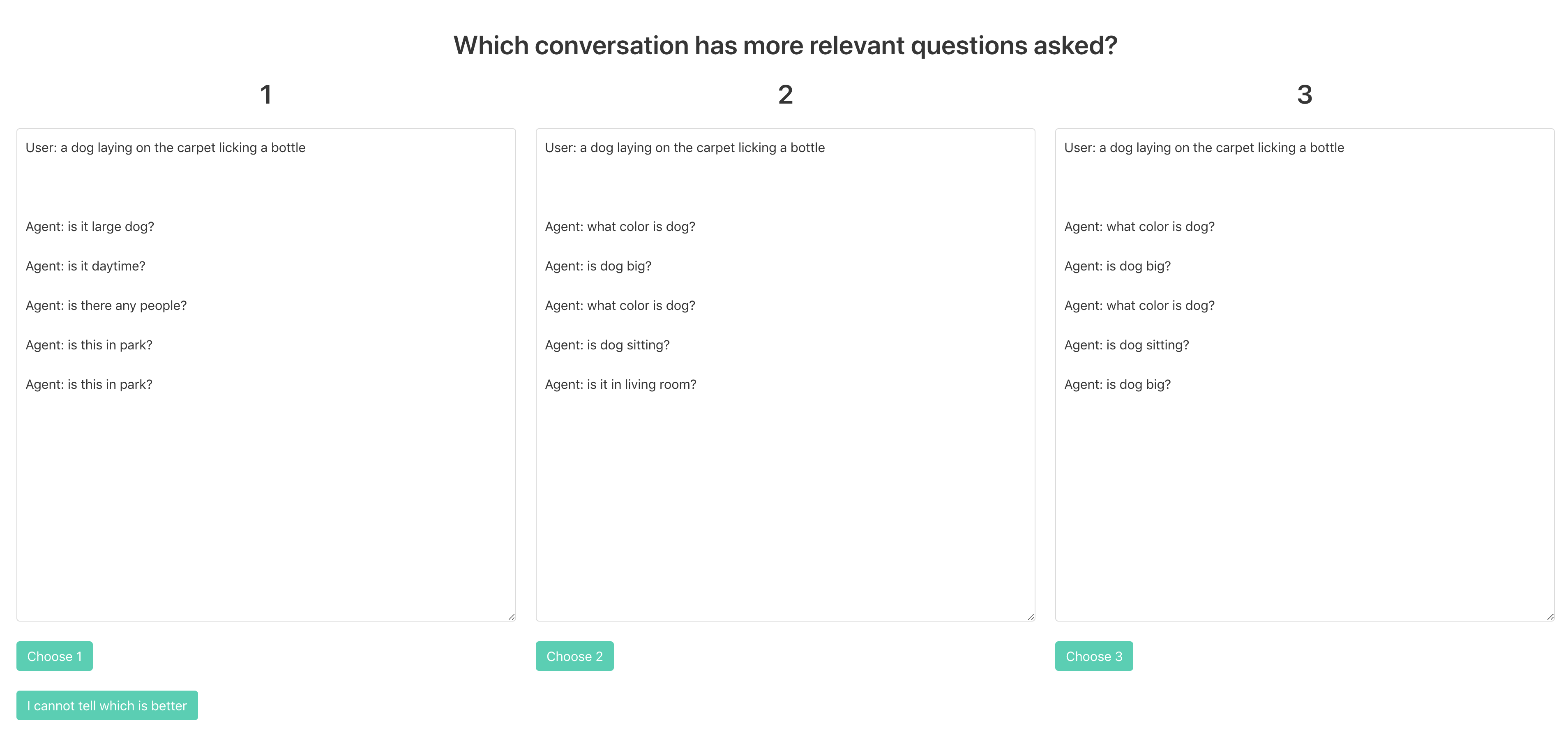}
\caption{The user interface for human relevancy experiment. The conversation from the three different models on the same game setting are presented in the three boxes. On top of every conversation is the caption for the target image. Followed by that are the five questions asked by the Q-Bot in that game.}
\label{fig:4}
\end{figure*}
\appendix
\section {Human Evaluation Interface}\label{appendix:interface}
We demonstrate our web interface for our Human-AI guessing game in Fig \ref{fig:3}. At the beginning of every game the human player is presented with a random sampled target image. For the trained Q-bot, 20 images are sampled as the candidate pool with the objective of retrieving the target image. At the end of the conversation, the Q-Bot's guess is revealed to the human player followed by a request to evaluate the quality of the conversation in the four aspects introduced in section \ref{human eval setting} on a scale from 1 to 5.

Our human relevancy evaluation interface is demonstrated in Fig \ref{fig:4}. We present side-by-side questions generated in the same game from each of the three different models and ask the human evaluators to pick the conversation that is most semantically relevant to the target image caption.  

\begin{table*}[!b]
\resizebox{\textwidth}{!}{%
\begin{tabular}{llll}
\hline Image & RL-Q-IG & SL-Q-IG & RL-Q \\ \hline \\ [-1.5ex]

\multirow{1}{*}{
\includegraphics[width=5cm]{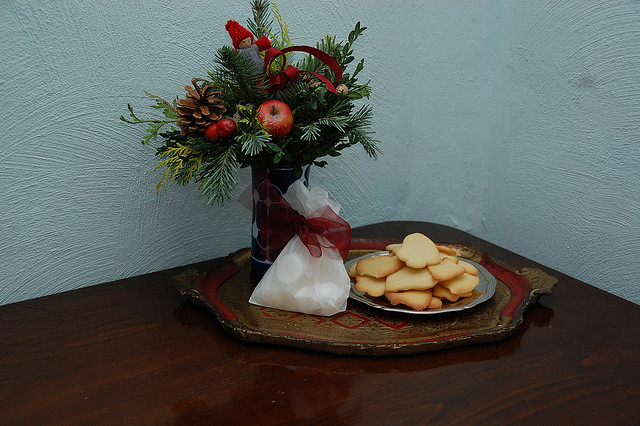}
}

& Q: \color{bleudefrance} what color is vase? & Q: \color{bleudefrance} what color is vase? & \color{carrotorange}Q: what color is table? \\
& A: it is transparent & A: it is brown & A: it is orange \\ 
& Q: \color{bleudefrance} is it on table? &  \color{carrotorange} Q: is it in vase? & \color{carrotorange}Q: is there anything else on table? \\ 
& A: yes it is & yes it is & A: yes there are flowers \\ 
& Q: \color{bleudefrance} is it in kitchen? & \color{carrotorange}Q: is it in vase? & \color{carrotorange}Q: what color is table? \\ 
& A: I do not know & A: yes it is & A: it is orange \\
& Q: any people? & \color{carrotorange}Q: is it in vase? & \color{carrotorange}Q: is there anything else on table? \\ 
& A: no, there is no people & A: yes it is & A: yes there are \\ 
& Q: \color{bleudefrance} what color is the table? & \color{carrotorange}Q: is it in vase? & \color{carrotorange}Q: is there anything else on table? \\ 
& A: it is brown & A: yes it is & A: yes there is \\ 

\end{tabular}}
\caption{A dialog example with the human generated caption: \textbf{there is a plant in a vase and cookies.} \color{bleudefrance} Blue \color{black} highlights diverse questions and \color{carrotorange} orange \color{black} indicates poor diversity.}
\label{tab:my_label}
\end{table*}

\begin{table*}[!b]
\resizebox{\textwidth}{!}{%
\begin{tabular}{llll}
\hline Image & RL-Q-IG & SL-Q-IG & RL-Q \\ \hline \\ [-1.5ex]

\multirow{1}{*}{
\includegraphics[width=5cm]{truck_image.jpg}
}
 
& Q: \color{bleudefrance} are men old?  & \color{bleudefrance} Q: how old are men? & \color{carrotorange} Q: what color is hat? \\
& A: No they are not & A: 30 years old & A: there is no hat \\ 
&\color{bleudefrance} Q: are they in uniform? & Q: are they in city? & Q: is it sunny? \\ 
& A: I'm not sure & A: yes they are & A: it seems yes \\ 
& Q: is it sunny? & Q: is it sunny? & Q: is this in city? \\ 
& A: yes it is & A: yes it is & A: it is in city \\
& \color{bleudefrance} Q: are they on road? & Q: any other people? & Q: are there any people in picture? \\ 
& A: yes they are & A: no other people & A: there are two people \\ 
& Q: \color{bleudefrance} are they in parking lot? & Q: animals? & \color{carrotorange} Q: is this in home or restaurant? \\ 
& A: No they're not & A: no other animals & A: it is outside \\ 

\end{tabular}}
\caption{A dialog example with the human generated caption:  \textbf{two men in formal wear standing next to a monster truck.}  \color{bleudefrance} Blue \color{black} highlights ideal relevant questions and \color{carrotorange} orange \color{black} indicates less relevant questions.}
\label{tab:my_label}
\end{table*}


\begin{table*}[!b]
\resizebox{\textwidth}{!}{%
\begin{tabular}{llll}
\hline Image & RL-Q-IG & SL-Q-IG & RL-Q \\ \hline \\ [-1.5ex]

\multirow{1}{*}{
\includegraphics[width=5cm]{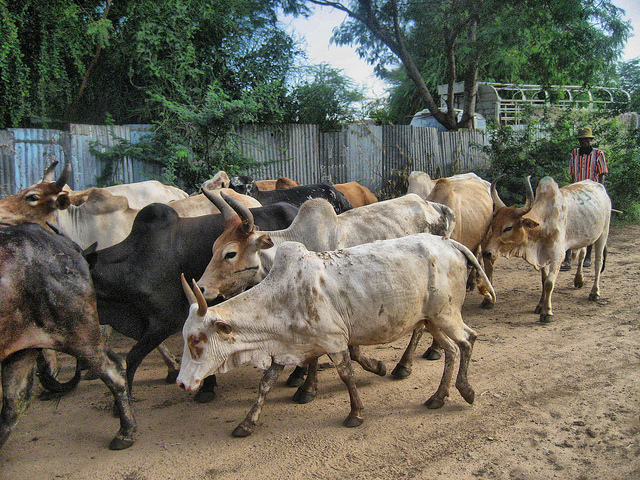}
}
 
& Q: \color{carrotorange} how many horses are there?  & Q: \color{carrotorange} how many horses are there? &  Q: is this in park? \\
& A: there are no horses & A: There are no horses, only bulls & A: maybe it is in park \\ 
& Q: \color{bleudefrance} are they in wild? & Q:  \color{carrotorange} what color are horses? & Q: \color{carrotorange} are there any people in picture? \\ 
& A: no, they are in village & A: they are white and black & A: yes there is one picture \\ 
& Q: \color{bleudefrance} is there fence? & Q: \color{bleudefrance} are they in field? & Q: \color{carrotorange} is this in home or restaurant? \\ 
& A: yes, there is fence & A: Yes they are & A: it is in neither of those \\

& Q: any people? & Q: \color{carrotorange} is it sunny? & Q: \color{carrotorange} are there any people in picture? \\ 
& A: yes, a person & A: Yes it is & A: yes there is one person \\ 
& Q: \color{carrotorange} animals? & Q: \color{carrotorange} is it sunny? & Q: \color{carrotorange} is this in restaurant? \\ 
& A: yes, several bulls & A: Yes it is & A: no it is not \\ 

\end{tabular}}
\caption{A dialog example with the human generated caption:  \textbf{an image of running with the bulls outside.}  \color{bleudefrance} Blue \color{black} indicates ideal relevant questions and \color{carrotorange} orange \color{black} highlights irrelevant/repeating questions.}
\label{tab:my_label}
\end{table*}

\begin{table*}[!b]
\resizebox{\textwidth}{!}{%
\begin{tabular}{llll}
\hline Image & RL-Q-IG & SL-Q-IG & RL-Q \\ \hline \\ [-1.5ex]

\multirow{1}{*}{
\includegraphics[width=5cm]{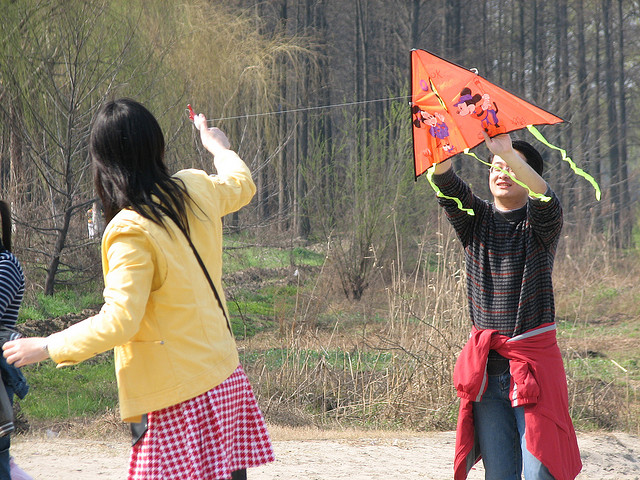}
}

& Q: \color{bleudefrance} what color is kite? & Q: \color{carrotorange} what color is horse? & Q: \color{carrotorange} is there only 1 person? \\
& A: It is orange & A: there is no horse & A: no there is not \\ 
& Q: is it sunny? & Q: \color{bleudefrance} how old is man? & Q: \color{bleudefrance} is this in park? \\ 
& A: Yes it is & A: he is young & A: yes it is \\ 
& Q: any other people? & Q: \color{bleudefrance} is he wearing hat? & Q: \color{carrotorange} are there any people in picture? \\ 
& A: Yes there are two people & A: no, he is not wearing hat & A: yes there are \\
& Q: any animals? & Q: \color{carrotorange} what color is horse? & Q: \color{carrotorange} is there any other food on table? \\ 
& A: No there is not & A: there is no horse & A: no food on the table \\ 
& Q: \color{bleudefrance} any trees? & Q: is it sunny? & Q: \color{carrotorange} is there anything else on table? \\ 
& A: Yes, there are several trees & A: yes it is & A: nothing . . \\ 

\end{tabular}}
\caption{A dialog example with the human generated caption: \textbf{a man holding a kite while a girl tries to fly it.} \color{bleudefrance} Blue \color{black} indicates ideal relevant questions and \color{carrotorange} orange \color{black} indicates poor relevance.}
\label{tab:my_label}
\end{table*}

\section{Qualitative Examples}\label{appendix:qualitative}

\end{document}